%% file: main-arxiv.tex
\documentclass[10pt,twocolumn,letterpaper]{article}

\usepackage{cvpr}
\input{preamble}
\usepackage[accsupp]{axessibility} 

\definecolor{cvprblue}{rgb}{0.21,0.49,0.74}
\usepackage[pagebackref,breaklinks,colorlinks,citecolor=cvprblue]{hyperref}

\def\confName{CVPR}
\def\confYear{2024}

\title{Learning from One Continuous Video Stream}

\author{Jo\~ao Carreira$^{1\dag}$, Michael King$^{1\dag}$, Viorica P\u atr\u aucean$^{1\dag}$, Dilara Gokay$^{1\dag}$, C\u at\u alin Ionescu$^{1\dag}$, \\ Yi Yang$^{\dag}$, Daniel Zoran$^{\dag}$, Joseph Heyward$^{\dag}$, Carl Doersch$^{\dag}$, Yusuf Aytar$^{\dag}$,\\ Dima Damen$^{\dag\ddag}$, Andrew Zisserman$^{\dag\Diamond}$\\
\small{$^{\dag}$Google DeepMind, $^{\ddag}$University of Bristol, $^{\Diamond}$University of Oxford}\\
\small{Corresponding author: \texttt{joaoluis@google.com}, $^{1}$core contributor}
}

\begin{document}
\maketitle
\input{sec/0_abstract}    
\input{sec/1_intro}
\input{sec/2_related}
\input{sec/3_method}
\input{sec/4_experiments}
\input{sec/5_conclusion}
{
    \small
    \bibliographystyle{ieeenat_fullname}
    \bibliography{main}
}

\input{sec/X_suppl}

\end{document}

%% file: preamble.tex
\usepackage[dvipsnames]{xcolor}

\newcommand{\TODO}[1]{\textbf{\color{red}[TODO: #1]}}

\renewcommand{\TODO}[1]{}

\usepackage{multirow}

%% file: sec/0_abstract.tex
\begin{abstract} We introduce a framework for online learning from a single continuous video stream -- the way people and animals learn, without mini-batches, data augmentation or shuffling. This poses great challenges given the high correlation between consecutive video frames and there is very little prior work on it. Our framework allows us to do a first deep dive into the topic and includes a collection of streams and tasks composed from two existing video datasets, plus methodology for performance evaluation that considers both adaptation and generalization. We employ pixel-to-pixel modelling as a practical and flexible way to switch between pre-training and single-stream evaluation as well as between arbitrary tasks, without ever requiring changes to models and always using the same pixel loss. Equipped with this framework we obtained large single-stream learning gains from pre-training with a novel family of future prediction tasks, found that momentum hurts, and that the pace of weight updates matters. The combination of these insights leads to matching the performance of IID learning with batch size 1, when using the same architecture  and without costly replay buffers. An overview of the paper is available online at ~\url{https://sites.google.com/view/one-stream-video}.
\end{abstract}

%% file: sec/1_intro.tex
\section{Introduction}

Humans gather knowledge about themselves and the world through a continuous stream of observations, from their early days as infants.
Some experiences are seen once in a lifetime while others are daily repeats.
With time, familiarity with environments, objects and experiences form a base to our knowledge, and memory maintains the most important experiences from our ever-further past.
Learning successfully \textit{adapts} to the current surroundings so we can make better predictions over time, while leading to \textit{generalization} to unseen environments.
Despite the naturalness of this approach of learning for humans, video understanding approaches have rarely attempted a similar regime. 

\begin{figure}[h!]
    \centering
    \includegraphics[width=\linewidth]{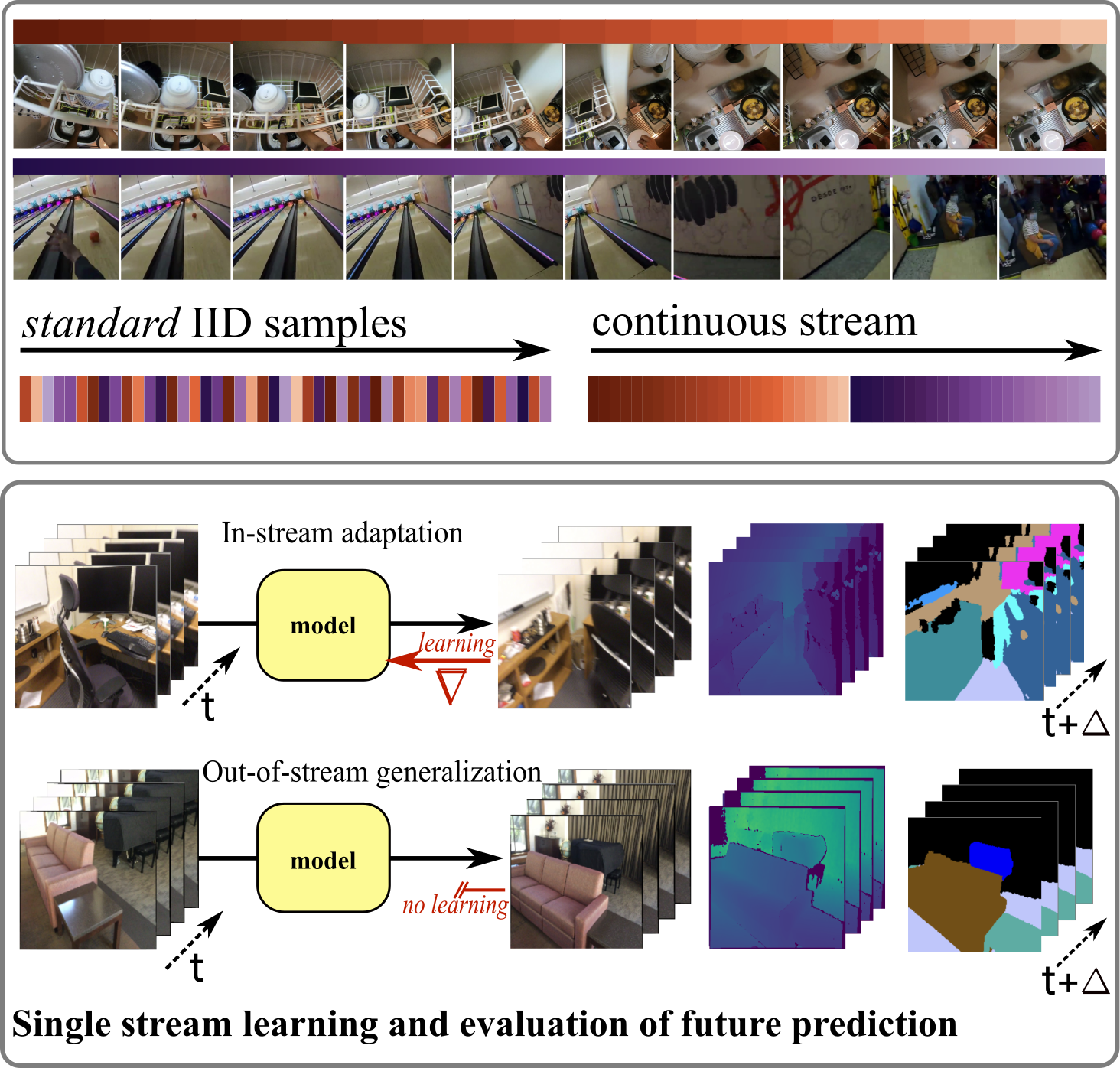}
    \caption{\textbf{Top:} We introduce a framework for studying continuous learning in a single video stream. This is a natural yet unstudied problem, different from standard independent and identically distributed (IID) learning in video where batches contain clips from random videos in a random order. \textbf{Bottom:} We propose pixel-to-pixel models to evaluate our approach across prediction tasks (prediction of future frames, depth, segmentation). We measure both adaptation to the video stream -- the model here updates its weights (learns) continuously to improve prediction -- as well as generalization to out-of-stream clips -- the model being adapted on the first stream is now evaluated on a different held-out stream without being allowed to adapt to it. We propose to maximize both adaptation and generalization.}
    \label{fig:teaser}
    \vspace{-3mm}
\end{figure}

The most related fields are continual and lifelong learning, but these are still deeply fractured -- no single problem formulation or benchmark is widely accepted. They also mostly focus on simplifications of the problem relative to human and animal experience, for example a popular task is learning one ImageNet class at a time and minimizing forgetting of previous classes~\cite{rusu2016progressive}. 

\begin{figure}
    \centering
    \includegraphics[width=.48\linewidth]{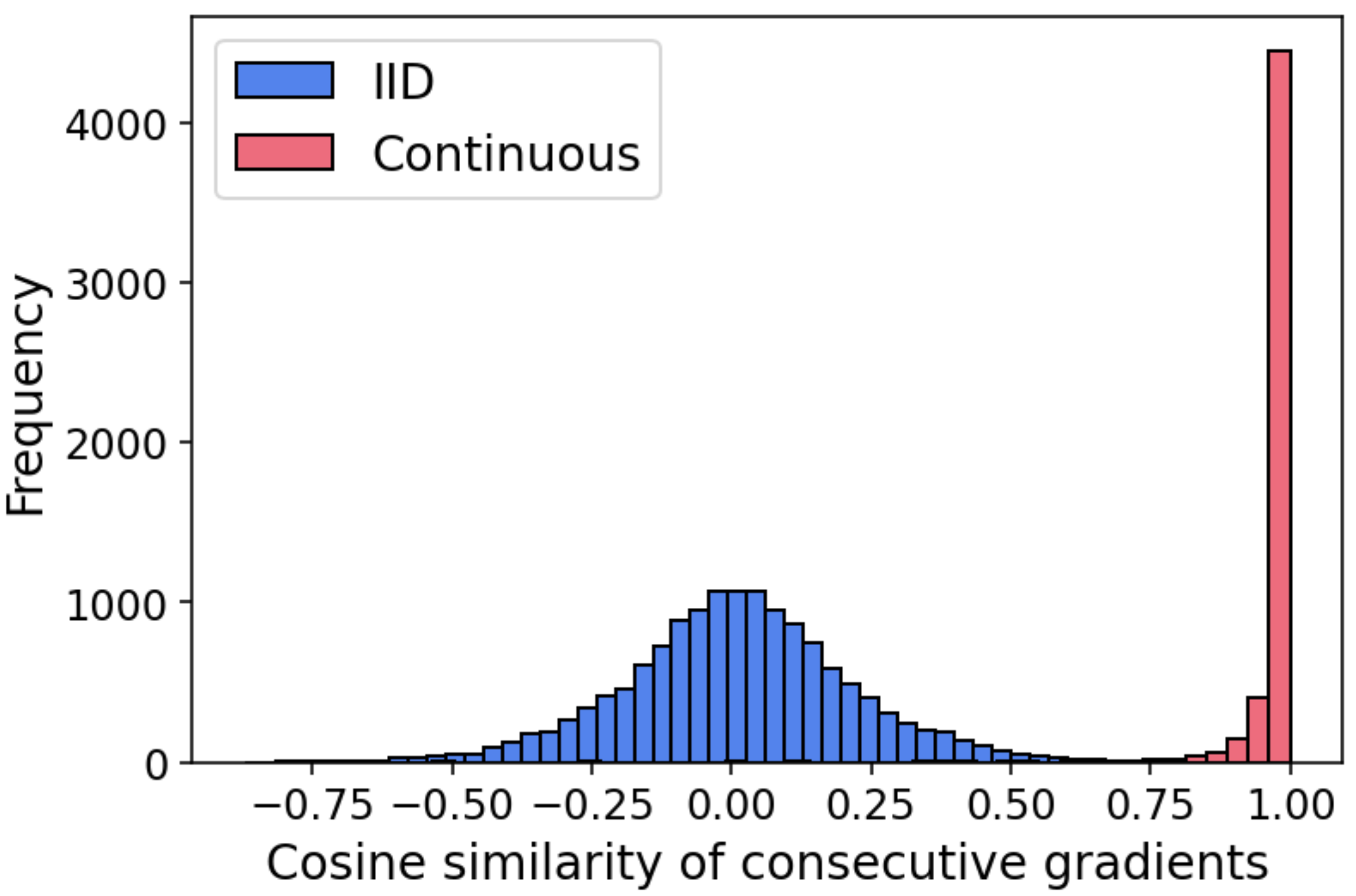}
    \includegraphics[width=.48\linewidth]{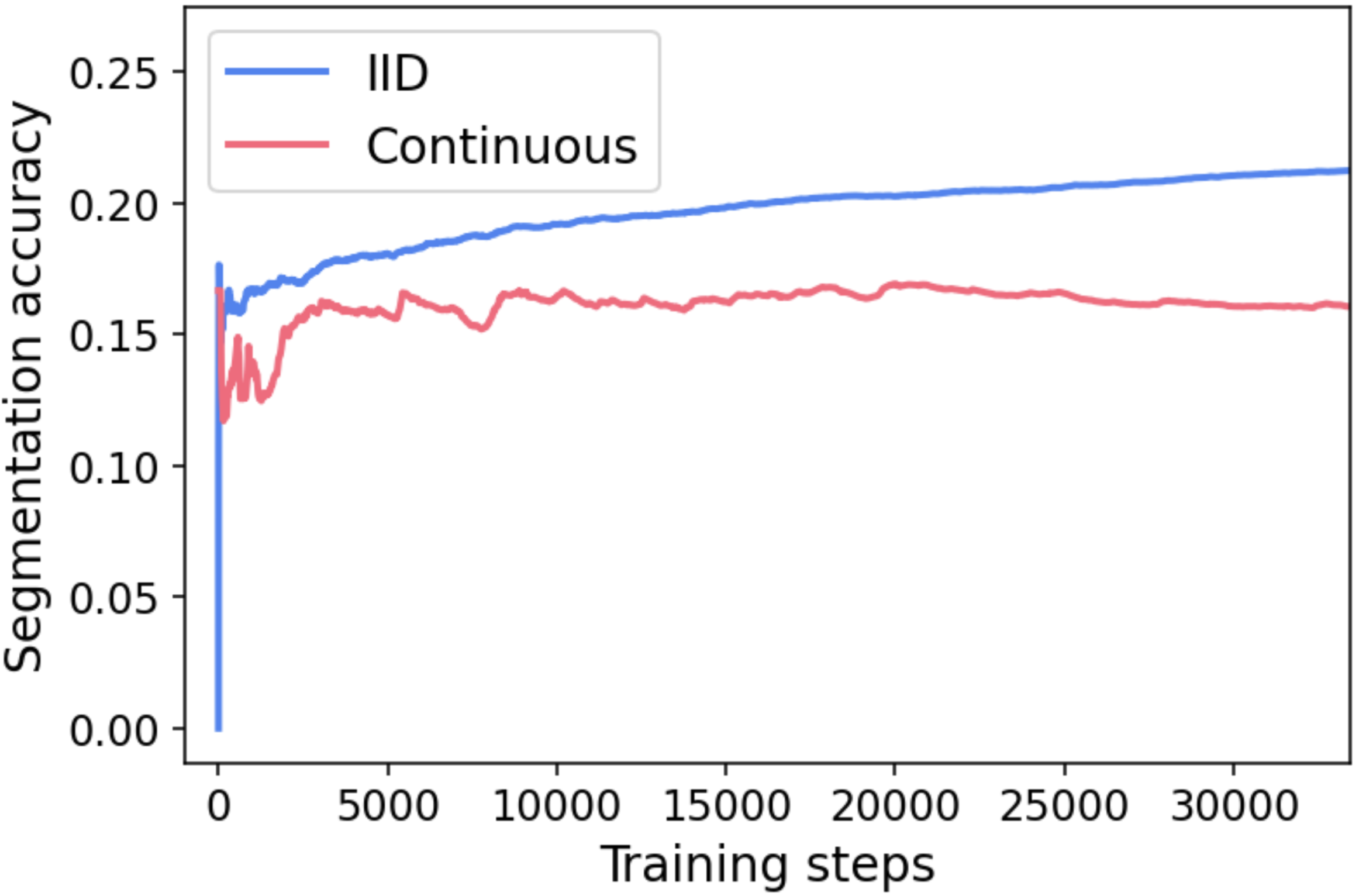}
    \vspace{-2mm}
    \caption{UNet training on ScanNet-stream for the semantic segmentation task. \textbf{Left}: The cosine similarity of consecutive gradients is normally distributed when training on IID data, but shows very strong correlations when training on a continuous video stream. \textbf{Right}: This is reflected in poor training performance. See the appendix for similar figures for Ego4D-stream.}
    \label{fig:cos_sim}
    \vspace{-4mm}
\end{figure}

But what happens if we attempt to learn continuously from a single video stream (Figure~\ref{fig:teaser}), meaning batch size 1 and high frame rate (\eg 25 fps)? We do not see thousands of dogs per second, unlike the standard ImageNet setting, where images of even a same dog breed are quite uncorrelated at several scales -- capturing a wide diversity of edges, textures, \etc. Standard deep learning approaches lead to smooth optimization there, but it is very different from the setting of learning from a single video stream, where things happen slowly, at much longer timescales, with great correlation or even no change between frames for periods of time; see Figure~\ref{fig:cos_sim}. Do standard deep learning tools work in this setting? This is currently unknown territory, with very few attempts touching it at all~\cite{contssl} -- and it is the focus of this paper.

We are interested in this problem because models should be able to adapt after deployment to their environment, hence receiving a single stream of information. This is needed for embodied intelligence (robotics), but also for any type of digital assistant that we would want to be able to adapt to a specific user's needs.

\vspace{1mm}
\noindent \textbf{A framework.} We lay out a framework for learning from a single video stream, focusing on the entry-level problem of ... learning. Nevermind forgetting, can the models even learn in the first place? We study how successful learning is for different optimization settings, models, pretraining methods and tasks, when done from a single continuous stream. Our methodology follows the following principles: 

\begin{itemize}
  \item We consider future prediction tasks including pixels-to-pixels, pixels-to-segmentation and pixels-to-depth, all in RGB space (\ie outputs are always 3 dimensional and with the same shape as the input frames). This way we do not need to change any parameter in the model  between pretraining and single-stream learning, independently of what tasks are used in each. It enables using a simple L2 loss between pixel values in all cases.
  \item We evaluate performance in two ways -  \textit{in-stream}, and \textit{out-of-stream}. In-stream   measures performance on the input video stream. For sufficiently difficult prediction tasks this should be enough. However, it conflates representation quality with adaptation to the scene, lighting condition, etc. For example if the lights in a scene are turned off for an hour, a model can learn to always predict black pixels -- which is not ideal. To disentangle adaptation and generalization, we also evaluate out-of-stream, on a second stream on which the model does not learn. We propose to maximize both.
\end{itemize}

\vspace{1mm}
\noindent \textbf{Contributions.} With this methodology in place, we identified the following new results:
\begin{itemize}
\item On the optimization front, we observe that momentum, widely used in popular optimizers such as Adam, is unhelpful in highly correlated video streams. Instead methods without momentum such as RMSprop are more robust. 
\item Another result is a trade-off between adaptation and generalization induced by frequency of weight updates -- slower leads to better generalization. 
\item We introduce a family of future prediction pretraining tasks, learnt in the standard independent and identically distributed (IID) setting, with large batches, and show that these lead to better single-stream performance compared to existing ImageNet pretraining tasks. 
\item We combine these results into a combo approach that we call \textit{Baby Learning} (BL), and compare it to a standard deep learning setup (STDL) that uses Adam and ImageNet MAE pretraining. We show that BL on a sequential stream matches STDL on out-of-stream generalization when that same stream is permuted to be IID, while achieving better performance in-stream thanks to adaptation.
\end{itemize}

\label{sec:intro}

%% file: sec/2_related.tex
\section{Related work}
\label{sec:related}

\vspace{1mm}
\noindent \textbf{Online learning from a single video stream.} One of the few papers dealing with this problem~\cite{contssl} proposed a minimum-redundancy ``replay'' buffer~\cite{mnih2015humanlevel} to deal with temporal correlations. Replay buffers are not the sexiest research avenue for continual learning~\cite{sutton:2022} and increase computation proportionally to the size of the buffer used. Additionally, in~\cite{contssl} performance on the video streams itself was never evaluated, only on unrelated image classification tasks. \text{Test-time training} is the next closest problem, such as presented in TTTVS~\cite{wang2023test} (also related ~\cite{wu2023labelefficient}). TTVS's motivation is to ``improve \textbf{inference} quality" by leveraging self-supervised test-time optimization. Our motivation is to propose single-stream \textbf{learning}, which is reflected in much longer streams (24h in this paper, compared to seconds or few minutes). One challenge in single-stream learning is how to fully exploit available hardware parallelism in a batch size 1 setting~\cite{carreira2018massively, malinowski2020sideways, malinowski2021gradient}.  

\vspace{1mm}
\noindent \textbf{Learning from a single video.} Also relevant are experiments training ConvNets~\cite{lecun1998gradient} or ViTs~\cite{dosovitskiy2020image} on a single image~\cite{Asano2020Single} or a long video~\cite{venkataramanan2023imagenet}. While outside of the streaming setting -- using data shuffling, augmentation and large batches, these papers showed that learning can be quite successful in this setting (especially for the shallower layers), which was quite encouraging.

\vspace{1mm}
\noindent \textbf{Continual learning.} The continual learning literature (\eg EWC ~\cite{Kirkpatrick2016OvercomingCF}) tends to focus on learning a sequence of tasks,  the goal being to learn new tasks as fast as possible without forgetting previous ones, such as in class incremental learning (\eg iCarl ~\cite{icarl}) and CLEAR~\cite{lin2021clear}. Most relevant is online continual learning, where models learn from a stream of data, visited once. One challenge is model cheating by exploiting labels in the data ~\cite{hammoud2023rapid}, or by learning spurious features~\cite{onpro} that have limited generalisation power. Most of the work in this area has been on collections of images, rather than video.

\vspace{1mm}
\noindent \textbf{Representation learning to the rescue.} Recent continual learning papers reported that some of the challenges of the problem are partially mitigated by using features pretrained in \textit{IID} settings~\cite{JMLR:v24:22-0496,ramasesh2022effect,wang2023comprehensive}. Various pretraining strategies exist: self-distillation (\eg BYOL ~\cite{byol}, BRaVe~\cite{brave}, DINO~\cite{dino}), contrastive (CLIP~\cite{clippmlr}, VideoMoco~\cite{pan2021videomoco}, VITO~\cite{parthasarathy2023self}), and masked auto-encoding (\eg VideoMAE~\cite{tong2022videomae}, TubeViT~\cite{tubevit}, AudioVisualMAE~\cite{Georgescu_2023_ICCV}, SiameseMAE~\cite{gupta2023siamese}, ~\cite{Bear2023UnifyingV}). We investigate this aspect in our setup.

%% file: sec/3_method.tex
\section{The Framework}
\label{sec:method}

\vspace{1mm}
\noindent \textbf{Overview.} The following is repeated throughout a video stream: a (potentially pretrained) model processes $n$ frames (a \textit{time step}) from an input video stream, and predicts frames for a future time step. Online (in-stream) performance metrics as well as an L2 loss function are then recorded by comparing predicted and  target frames. Given the loss, gradients are computed and the model  weights are updated by an optimizer. The target frames can come from the input stream itself or from an associated stream (e.g. for datasets having semantic segmentation or depth). In parallel with in-stream evaluation, we periodically assess performance on a held-out stream (composed of clips from the validation set of the dataset), to assess \textit{generalization}.

\subsection{Unified pixel-to-pixel modelling}

We are interested in having a framework where switching between arbitrary tasks requires no changes to models or losses so we can abstract away  decoder nd loss function design (large research areas) and focus on the single-stream learning aspect. To achieve this we map all task target outputs to RGB space, so we can have a single pixel-to-pixel model and a single simple L2 pixel loss. 

\vspace{1mm}
\noindent \textbf{Models.} We use 2 different popular backbones for our experiments: a UNet with self-attention in the bottleneck layer~\cite{NEURIPS2020_4c5bcfec} and ViT-L~\cite{dosovitskiy2020image}. To enable motion understanding, which is important for future prediction, we feed the models $n=4$ consecutive frames stacked along the channel dimension. We also have the models predict the same number of frames, so 12 channels in total (4 frames times 3 channels, for RGB). For ViT we decode tokens to pixels using a channel to space transformation: each token gets mapped to an appropriate number of channels using a linear transformation, then gets reshaped into a patch (e.g. a 16x16 grid of 4x3 RGB channels). We provide details of the models in the Appendix, but it is worth noting that the UNet is considerably smaller at 8M parameters, compared to the 350M parameters of the ViT model. We always train the UNet from scratch, whereas we explore various pretraining schemes for ViT. To accommodate for the 12 input channels we use, we inflate~\cite{carreira2017quo} the first layer of the ViT (replicate the pretrained weights 4 times).

\vspace{1mm}
\noindent \textbf{Memory.} We did not explore explicit memory modules such as memory banks, LSTM cells, or long context in this paper, but we do think it should be very relevant going forward. In all cases, the model ``sees'' only a time step of frames and predicts another time step. Note however that the model weights are updated as it goes through the video stream -- so both the weights and the optimizer (when stateful) provide some memory effects.

\noindent \textbf{Replay buffer.} We explored replay buffers~\cite{mnih2015humanlevel}, which keep a large cache of previous examples, then form batches by sampling from it. They have the disadvantage of increasing computational cost significantly over operating with batch size 1 (increases the theoretical computational cost of learning $K$ times, where $K$ is the batch size).

\subsection{Video streams and tasks}

\begin{table*}
    \centering
    \small{
    \begin{tabular}{l|cccccc}
    \hline
    Stream name & \# videos train & \# frames train &\# videos val &\# frames val & Max. length & Median length \\
    \hline
        Ego4D-stream & 21,704 & 294M (3,265h) & 2302 & 31M (348h) & 1.95h & 8.8 minutes \\
       ScanNet-stream & 1,199 & 1.8M (20h) & 312 & 0.5M (5.7h) & 5.5 minutes & 1 minute \\
        \hline
    \end{tabular}}
    \vspace{-2mm}
    \caption{The two streams we consider, formed out of Ego4D and ScanNet, together with properties of their original video clips.}
    \label{tab:streams}
\end{table*}

We do not know of public datasets having very long video streams, for example days-long and beyond\footnote{Other than Krishnacam~\cite{krishna-wacv2016}, which is only 70 hours long.}, so we create two different video streams: Ego4D-stream and  ScanNet-stream, by concatenating videos from, respectively, Ego4D and ScanNet. We include Ego4D as it is large and has very long videos, and ScanNet because it has dense semantic segmentation and depth annotations, allowing us to experiment with different tasks.

\vspace{1mm}
\noindent \textbf{Ego4D-stream.} We concatenate the raw (un-trimmed) videos from the Ego4D dataset~\cite{Ego4D2022CVPR} to create a very long video stream. We use $\sim$90\% of the data to generate a training stream and $\sim$10\% for the validation stream. 
The videos in Ego4D were collected using a head or glass mounted camera, capturing activities of daily life. We use this stream for learning and evaluating future prediction in pixel space -- no annotations are employed. See Table~\ref{tab:streams} for more details.

\vspace{1mm}
\noindent \textbf{ScanNet-stream.} We use the videos from ScanNetV2~\cite{dai2017scannet} to define 3 different prediction tasks, of pixels, semantic segmentation labels (40 classes) and  depth. The availability of these additional target streams makes it possible for us to measure more explicitly higher-level understanding, which may not be clear from pixel prediction alone. The videos in this dataset were filmed with a camera navigating indoor scenes with one or more rooms. See Table~\ref{tab:streams} for more details.

\vspace{1mm}
\noindent \textbf{Tasks.} Overall we consider the following tasks across the two stream types: Ego4D-stream pixel prediction, ScanNet-stream pixel, semantic segmentation and depth prediction. The difficulty of most of these tasks can be controlled using a time displacement parameter $\Delta$ -- how many steps in advance should the models predict. We consider 0, 1, and 4 time steps. A displacement of 0 only makes sense for semantic tasks as otherwise it corresponds to auto-encoding, which is easy. Note that this task is related to the near-future accuracy metric proposed in~\cite{hammoud2023rapid} to measure (forward transfer) adaptation in a more robust way.

\vspace{1mm}
\noindent \textbf{RGB-fication.} There is some hand design involved in mapping tasks into RGB space. For depth and semantic segmentation we use standard color maps typical for those tasks in ScanNet: the Viridis color map and the ScanNet label color map, respectively. We provide details of these mappings in the Appendix. For computing metrics other than the L2 pixel distance, we compute the nearest neighbor label / depth for each pixel color (e.g. a pixel's color may not correspond exactly to one of the semantic segmentation labels, so we assume the predicted class is the label associated with the nearest color on the color map).

\subsection{Evaluation}

Learning a video stream task with a continuously updated model can be seen to decompose into: 1) discovering strong features that are general, for example learning that certain configuration of edges correspond to chair legs for semantic segmentation (\textbf{generalization}), which is likely to happen over long timescales. 2) by specializing to the particular scene being seen in the video, e.g. learning that the chair is the only red object in the scene and  using redness to label the chair pixels (\textbf{adaptation}), which can happen over short timescales. Adaptation can be quite useful and lead to efficient visual mechanisms, but may not necessarily lead to generalization. Dummy models that only \eg exploit temporal correlations in the target stream could misleadingly appear to perform very well~\cite{hammoud2023rapid}; see Table~\ref{tab:main_table}, \textit{Blind} model.  

We propose to evaluate both aspects by computing a score continuously \textbf{in-stream}, on the video stream task the model is learning from, to measure adaptation. Separately and periodically we compute the same score \textbf{out-of-stream}, on a held-out stream with the optimizer disabled (\textit{out-of-stream}). We use the training sets of Ego4D and ScanNet to do in-stream evaluation, and the validation sets for out-of-stream evaluation.

\vspace{1mm}
\noindent \textbf{Cumulative scores.} Comparing single-stream learning models and approaches can be difficult if done naively. On the training stream, difficulty of the task may vary over time, for example certain parts of a stream may be more unpredictable due to camera motion or other factors, leading to natural oscillations. On the generalization side we care not just about performance at a particular moment in time, but also how fast it ramps up as training on the sequential video stream progresses. With this in mind, we use a global score over the whole stream -- we do this by averaging the performance over 10,000 evenly-spaced points interpolated from the steps in both in-stream and out-of-stream settings.

\begin{figure*}[t]
\begin{center}
    \includegraphics[width=1.0\linewidth]{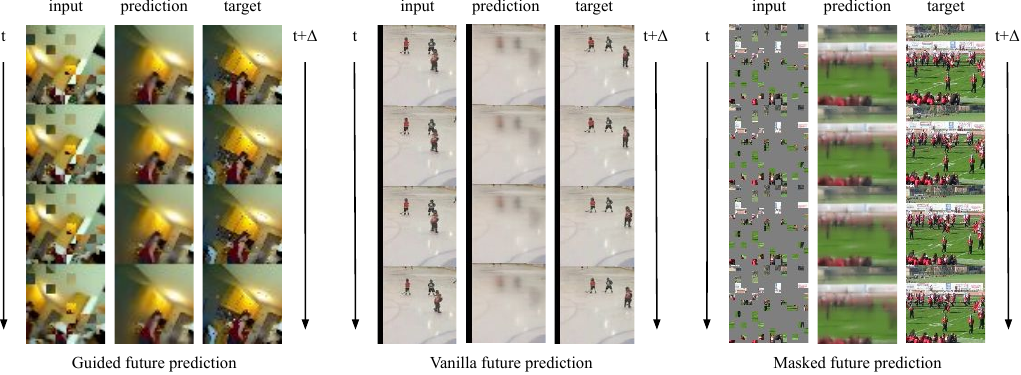}
\end{center}
\vspace{-5mm}
    \caption{Video pretraining tasks we consider, sorted from easiest to hardest, left to right -- guided future prediction, vanilla future prediction, and masked future prediction. Each column shows 4 consecutive frames vertically. For each method we show \textbf{left-to-right}: input frames, predictions from the model, target frames. We use a displacement ($\Delta$) of 16 frames (0.64s) between input and target clips.}
    \label{fig:vit-variations}
    \vspace{-0.3cm}
\end{figure*}

\vspace{1mm}
\noindent \textbf{Task-specific evaluation.} We evaluate pixel prediction with average L2 pixelwise distance, the same that we use as loss for training. For semantic segmentation we use mean per-frame IoU and recall~\cite{Pont-Tuset_arXiv_2017, Caelles_arXiv_2019} and depth using log relative mean square error (logRMSE)~\cite{eigen2014depth}. For both segmentation and depth we mask out pixels that were not annotated from the evaluation and loss. 

\section{Generalized Future Prediction}
\label{sec:future_pred}

One of the most promising directions  for continual learning, as identified in prior work, is representation learning. We are particularly interested in pixel-to-pixel approaches keeping consistent with the overall proposed framework where we never change neither the architecture nor the loss. We propose a family of pretraining methods that generalize future video prediction~\cite{srivastava2015unsupervised}, and that follow a same pattern: given one input video clip the model is trained to predict a future clip from the same video (both clips are 4 frames long). We consider three variants, illustrated in Fig.~\ref{fig:vit-variations}: 

\begin{itemize}
    \item The easiest one, \textit{Guided future prediction}, replaces a few patches from the input clip with patches in the same position from the future clip, hence narrowing down the range of possibilities.
    \item The intermediate one, \textit{Vanilla future prediction} is just the standard task of predicting the future.
    \item \textit{Masked future prediction} is a variation of Masked Auto-Encoding where the model must predict the future clip based on a partial view of the current clip.
\end{itemize}

Guided future prediction is related to Siamese MAE~\cite{gupta2023siamese, Bear2023UnifyingV, weinzaepfel2022croco}, but simpler, requiring a single forward pass over one model. Masked future prediction is related to video MAE-type approaches~\cite{tong2022videomae, feichtenhofer2022masked} but with a strict separation between disjoint input and output clips, whereas in traditional video MAE there is a single clip that gets  uniformly corrupted as part of denoising auto-encoding.

\vspace{1mm}
\noindent \textbf{Hyperparameters.} There are two hyperparameters: the fraction of guiding patches / masked patches and the shape of these patches (we use a constant square shape in all experiments). Vanilla future prediction is a special case of the other two in the case where there is no masked nor guiding patches. A general way to increase the difficulty of any of the variants is to increase the displacement ($\Delta$) between the input and the target clips.

%% file: sec/4_experiments.tex
\section{Results}

Our goal is to achieve similar (or better) learning efficiency from a single continuous video stream as we expect from the standard deep learning setting -- using sequences of batches of well shuffled examples. Here is a summary of the main results of our experiments on learning from a continuous video stream:

\begin{itemize}
\item  Momentum, widely used in optimizers such as Adam, hurts performance in single-stream learning.
\item  Less frequent weight updates (e.g. every 2.5 seconds), helps generalization while sacrificing some adaptation.
\item  Pretraining the models on IID data before single-stream learning is quite impactful. While popular ImageNet-based pretraining helps, we found  future-prediction based video pretraining to be vastly superior.
\end{itemize}

We ran all in-stream learning for
24 hours of video, and measured out-of-stream performance from regularly saved checkpoints on 3h30 of video. Resolution was always 224x224.

\subsection{Optimization}

\begin{figure}
    \centering
    \includegraphics[width=\linewidth]{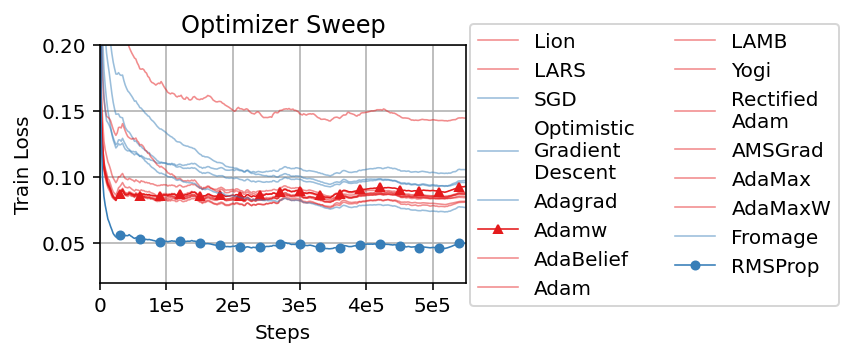}
    \vspace{-8mm}
    \caption{A sweep over commonly used optimizers. Those without momentum are shown in blue and aid the models adaptability considerably compared to the more commonly used Adam variants, which are shown in red.}
    \label{fig:optimizer_sweep}
    \vspace{-3mm}
\end{figure}

Starting from the standard optimizer used in nearly all deep learning setups, Adam, we observed that it does not train well with default parameters in the continuous video stream setting compared to the IID setting; see Fig.~\ref{fig:cos_sim}, right. To understand the differences between the two settings, we analyzed the temporal correlation between consecutive gradients. We found that the norm and variance of the gradients do not reveal strong differences. However, their orientation reveals strong correlations between pairs of consecutive gradients in the continuous case; see Fig.~\ref{fig:cos_sim}, left, where we plot the distribution of the cosine similarity between consecutive gradients in the continuous vs IID case. 

To investigate the optimization further we performed a large sweep over commonly used optimizers, shown in Fig.~\ref{fig:optimizer_sweep}. Results are averaged over 8 settings: 2 tasks (Ego4D-stream  and ScanNet-stream segmentation prediction), 2 models (ViT and UNet) and 2 displacements (1 step and 4 steps). RMS Prop significantly outperformed the more commonly used Adam variants.

\begin{figure}
    \centering
    \includegraphics[width=\linewidth]{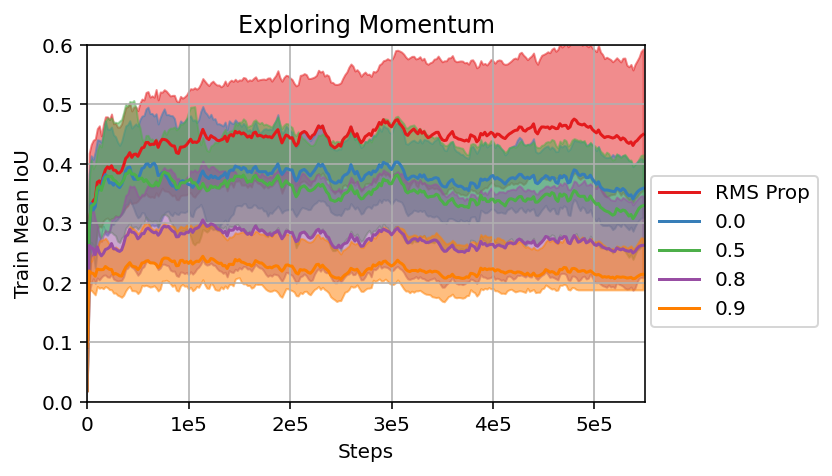}
    \vspace{-8mm}
    \caption{Reducing momentum with the AdamW optimizer helps to recover some of the performance of RMSProp.}
    \label{fig:momentum}
\end{figure}

\vspace{1mm}
\noindent \textbf{Momentum hurts.} To  scrutinize the difference between AdamW and RMSProp, we looked at the impact of \textit{momentum}. We found that lowering the momentum of the AdamW optimizer recovered some of the performance of RMSProp (as shown in Fig.~\ref{fig:momentum}). One possible explanation is that momentum exacerbates the problem of correlated consecutive gradients (that differ from the underlying gradient of the loss function over the whole stream) and makes the weights accelerate too much in the wrong direction. We stick with RMSProp for the rest of the paper, as it is more memory efficient due to not using the average of past gradients.

\vspace{1mm}
\noindent \textbf{Infrequent weight updates help generalization, hurt adaptation.} We also observed an interesting effect associated to the frequency of weight updates: doing it less frequently tends to benefit generalization at some cost to adaptation (since the model cannot adapt as frequently or as strongly to any particular time step). We found that updating weights every 16 frames (or 0.64s) provided a decent trade-off between adaptation and generalization across models, tasks and datasets as shown in Tab.~\ref{tab:freq_updates}

\begin{figure}
    \centering
    \includegraphics[width=0.9\linewidth]{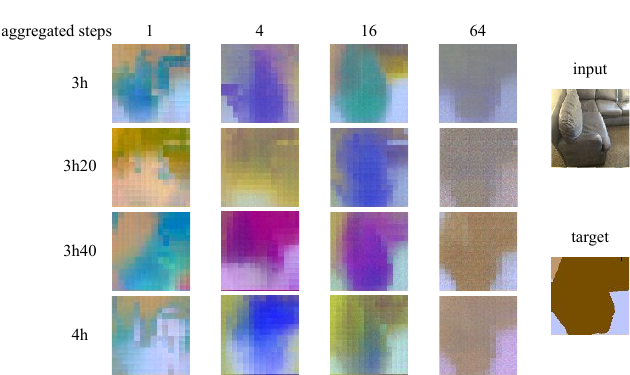}
    \caption{ScanNet-stream segmentation task \textit{generalization} results  on the exact same out-of-stream clip after different number of hours of exposure to a same stream for 4  model runs. The models differ only by the length of the interval between weight updates / number of steps of gradient aggregation (horizontal axis). Models with less frequent updates tend to generalize better (rightmost column), whereas models with more frequent updates tend to have strong priors about which objects are currently in the scene, leading to hallucinations (leftmost 3 columns in this case).}
    \label{fig:freq_updates_seg}
    \vspace{-3mm}
\end{figure}

\begin{figure}
    \centering
    \includegraphics[width=\linewidth]{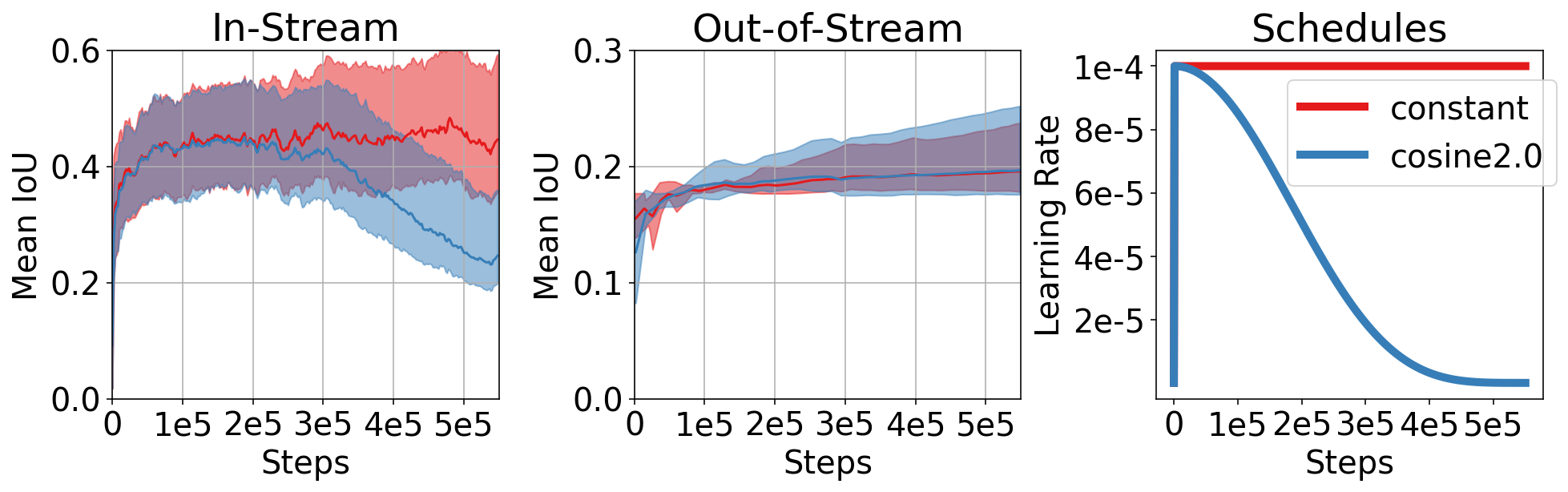}
    \vspace{-6mm}
    \caption{In-stream and out-of-stream results for a learning rate schedule using cosine decay with exponent 2.0 vs a constant learning rate (both with a linear warmup phase of 1k steps. The decaying learning rate is able to generalise more quickly but at a significant cost to adaptation.}
    \label{fig:lr_sched_sweep}
    \vspace{-2mm}
\end{figure}

\vspace{1mm}
\noindent \textbf{Constant learning rate helps adaptation.} We tested a range of common learning rate schedules including constant, linear, cosine and exponential decay, ``1cycle'' and cosine decay with restarts. The main finding was that decaying learning rates over the course of training, in particular cosine decay with exponent 2.0, helps generalization but significantly hurts adaptation as shown in Fig.~\ref{fig:lr_sched_sweep}. We therefore used a constant learning rate, with a linear warmup period of 1k steps, in our experiments.

\vspace{2mm}
\noindent \textbf{Replay buffer batches can be small.} We focus most of our exploration on the case without replay buffers, but did check what happens when we add a replay buffer to our very best setting - which slows the experiment down proportionally to the batch size employed. Over 11 hours of wall-clock training on the ScanNet-stream segmentation task with prediction displacement of 1 time step, an experiment with a replay buffer containing 10,000 samples (circular writing, random reading) and batch size 4 reached  mean IoU only 2\% superior to a model trained without replay buffer. Batch size 16 did not improve over batch size 4 and we did not use replay buffers in any other experiments.

\begin{table}
\centering
\small{
\begin{tabular}{c|c c c c c c}
\multicolumn{3}{c}{} & \multicolumn{4}{|c}{n steps per update} \\
\hline
Stream & dataset & model & 1 & 4 & 16 & 64 \\
\hline
\multirow{4}{*}{In} & \multirow{2}{*}{Ego4D ($\downarrow$)} & UNet & \textbf{.036} & .038 & .037 & .039 \\
 & & ViT & \textbf{.035} & .037 & .039 & .047 \\ \cline{2-7}
 & \multirow{2}{*}{Segm ($\uparrow$)} & UNet & \textbf{.420} & .292 & .211 & .195 \\
 & & ViT & \textbf{.457} & .395 & .302 & .232 \\
 \hline
 \multirow{4}{*}{Out-of} & \multirow{2}{*}{Ego4D ($\downarrow$)} & UNet & .095 & .051 & .047 & \textbf{.042} \\
 & & ViT & .076 & .062 & .046 & \textbf{.044}  \\ \cline{2-7}
 & \multirow{2}{*}{Segm ($\uparrow$)} & UNet & .176 & .179 & \textbf{.205} & .183 \\
 & & ViT & .251 & .272 & \textbf{.280} & .274 \\
 \hline
\end{tabular}}
\vspace{-3mm}
\caption{\label{tab:freq_updates}Results on video streams from two datasets, two models, different numbers of steps per gradient update (horizontal), and evaluated in-stream and out-of-stream, showing the impact of accumulating gradients over multiple steps. Results are averaged over 1 and 4 displacement steps. For Ego4D-stream we report average L2 pixelwise distance (lower is better), for ScanNet-stream Segm we report mean per-frame IoU (higher is better). Bold marks the best result for each model-stream pair.}
\vspace{-3mm}
\end{table}

\subsection{Pretraining}

\begin{figure}
    \centering
    \includegraphics[width=\linewidth]{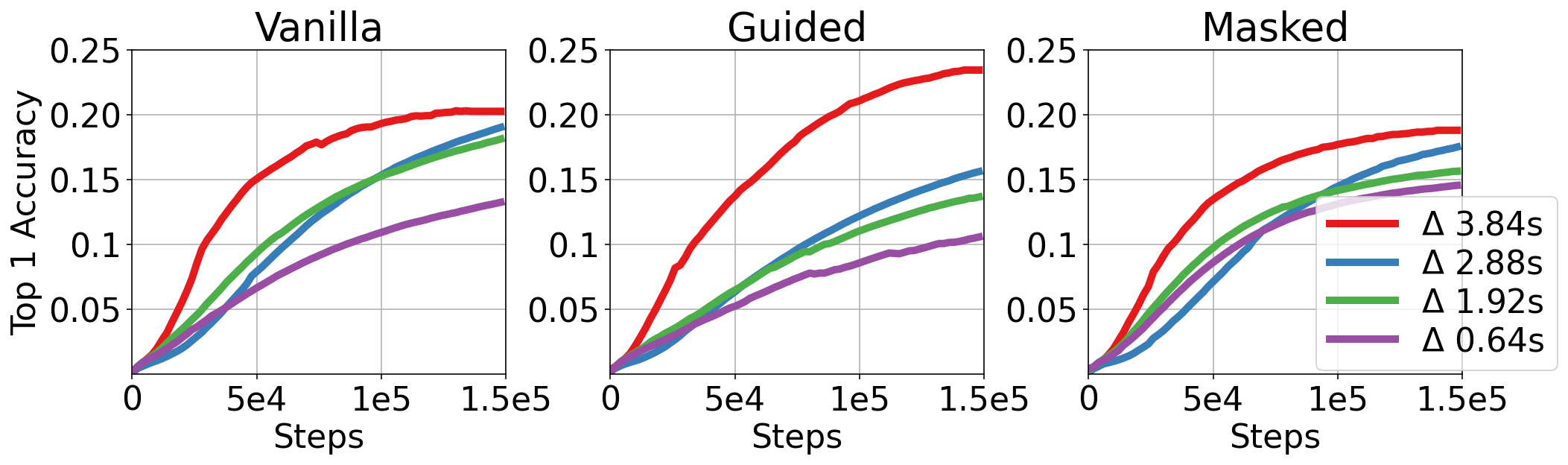}
    \vspace{-6mm}
    \caption{Linear top-1 accuracy in Kinetics during pretraining for different forms of future prediction tasks: Vanilla, Guided, Masked, and different displacements. The longer the displacement the better the accuracy is. Note that this is an online evaluation where the model only sees 4 frames (that for "Masked" and "Guided" are distorted) hence performance is far below models evaluated on the whole 10s of video (250 frames), often with plenty of test time augmentation, but the relative performance between the various curves is informative.}
    \label{fig:pretraining}
    \vspace{-4mm}
\end{figure}

For pretraining we did not learn from a single continuous stream, but instead formed big well-shuffled batches of 1024 clips using 8x8 slices of TPU-v5-lite, used AdamW and updated weights every step, for 150k steps. We pretrained on the Kinetics-700-2020 dataset~\cite{smaira2020short}, which is composed of 10s clips. We found that initializing with the ImageNet-MAE checkpoint led to much quicker optimization, so used that in all cases. More details can be found in the Appendix.

We pretrained ViT-L models on the future prediction tasks from sec.~\ref{sec:future_pred} and monitored their representation learning performance during training by adding a linear head to the model just before the decoder (we use a stack of 4 self-attention layers on top of the ViT-L  encoder). For this monitoring, we employed a standard cross-entropy loss between logits and ground truth labels in conjunction with a stop-gradient so the supervision would not influence the backbone weights.

We experimented with different input-target clip temporal displacements and observed the interesting effect that the longer the displacement the better the classification performance is as shown in Fig.~\ref{fig:pretraining}, even for masked future prediction, which reduces to the popular Masked Auto-encoding~\cite{he2022masked} when displacement is 0.  Our experiments suggest that 0  may be a sub-optimal choice. For the longest displacement the best method for top-1 accuracy emerged as Guided Future Prediction. The top result shown in the plot also used fewer guiding patches, 5\% instead of 10\% -- we used this best model for the rest of our evaluations. We show example video predictions for the various models online at~ \url{https://sites.google.com/view/one-stream-video}.

\noindent Table~\ref{tab:pretraining} has results on in/out-of-stream evaluation, showing large benefits over ImageNet classification pretraining and using no pretraining (training from scratch). ImageNet-MAE also does considerably better than classification-based pretraining. The same ViT-L architecture was used in all cases.

\begin{table*}
    \centering
    \small{
    \begin{tabular}{l|cccc}
    \hline
    Pretraining Checkpoint & Ego4D ($\downarrow$) & ScanNet Depth ($\downarrow$) & ScanNet Segm ($\uparrow$) & ScanNet ($\downarrow$) \\
    \hline
    None & .074 / .105 & 1.969 / 2.163 & .177 / .188 & .083 / .083 \\
    \hline
    ViT-L-I1K-CLS & .043 / .048 & 1.821 / 2.040 & .288 / .234 & .040 / .042 \\
    ViT-L-I21K-CLS & .042 / .048 & 1.735 / 2.013 & .244 / .192 & .039 / .040 \\
    ViT-L-I1K-MAE & .040 / .044 & 1.806 / 2.045 & .360 / \textbf{.320} & .037 / .038 \\
    \hline
    Guided Future Prediction & \textbf{.036} / \textbf{.043} & \textbf{1.622} / \textbf{1.990} & \textbf{.390} / .313 & \textbf{.032} / \textbf{.034} \\
    \hline

    \end{tabular}}
    \vspace{-3mm}
    \caption{\label{tab:pretraining} Table comparing performance of models pretrained on ImageNet-1K and 21K using MAE or classification, and models pretrained on our video tasks in Kinetics. The results are averaged over two sets of experiments, with displacements of 1 and 4 time steps and are shown in the format in-stream / out-of-stream.}
    \label{tab:your_label}
    \vspace{-3mm}
\end{table*}

\begin{figure*}[h]
    \centering
    \includegraphics[width=.85\linewidth]{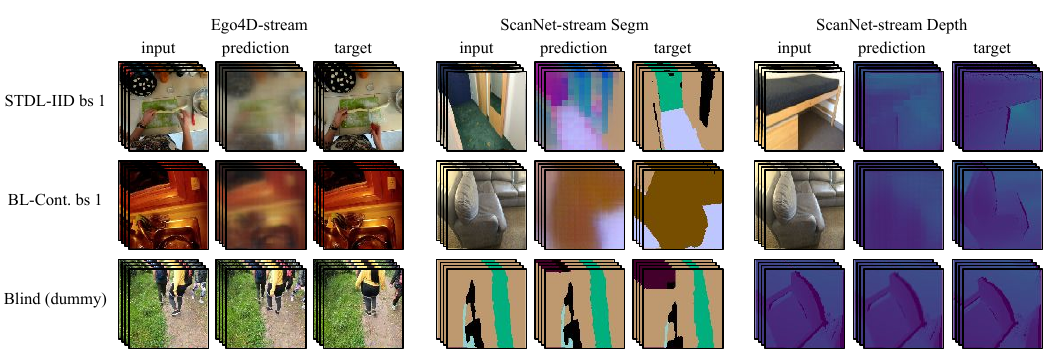}
    \vspace{-2mm}
    \caption{Future prediction results for video streams from Ego4D-stream, ScanNet-stream Segm, and ScanNet-stream Depth.}
    \label{fig:main_fig}
    \vspace{-1mm}
\end{figure*}

\begin{table*}[h!]
\centering
\small{
\begin{tabular}{c|c c | c c | c c | c c}
& \multicolumn{2}{|c|}{Ego4D ($\downarrow$)} & \multicolumn{2}{c|}{ScanNet ($\downarrow$)} & \multicolumn{2}{c|}{ScanNet Segm ($\uparrow$)} & \multicolumn{2}{c}{ScanNet Depth ($\downarrow$)} \\
\hline
displacement ($\Delta$) & t+1 & t+4 & 
t+1 & t+4 & 
t+1 & t+4 & 
t+1 & t+4  \\
\hline
STDL (IID) bs 16 & .023/.019 & .060/\.055 & .014/\.013 & .042/.061 & .495/.398 & .382/.295 & 1.798/2.01 & 1.838/2.038 \\
\hline
STDL (IID) bs 1 & .019/\textbf{.018} & .057/\textbf{.056} & \textbf{.010}/.013 & \textbf{.051}/\textbf{.060} & .376/.302 & .276/.227 & 1.722/\textbf{2.012} & 1.759/\textbf{2.034} \\
BL (Cont.) bs 1 & \textbf{.018}/.021 & \textbf{.055}/.066 & .011/\textbf{.012} & .055/.061 & \textbf{.463}/\textbf{.312} & \textbf{.328}/\textbf{.241} & \textbf{1.595}/2.038 & \textbf{1.655}/2.097 \\
\hline
Blind (dummy) & .038 / - & .086 / - & .032/ - & .109/ - & .547/ - & .307/ - & 1.256 / - & 1.649 / - \\
\hline
\end{tabular}}
\caption{\label{tab:main_table}Future prediction results for video streams from various datasets for two different temporal displacements (horizontal). We show results for the standard deep learning approach (STDL) on IID data with batch size 16 and 1, and for our approach (BL) when using a continuous video stream (Cont.). Two numbers are reported for every cell, corresponding to in-stream / out-of-stream performance. We highlight which of the two batch size 1 approaches performs best for each number. The Blind (dummy) model exploits correlations in the target stream which is available in-stream and used for learning, hence performs very well in-stream, but has random performance in evaluation where the target stream is not available. This shows the necessity to measure both adaptation and generalization performance.}
\end{table*}

\subsection{IID vs Continuous Learning}

We compare our best setup, which we call, for no particularly technical reason, ``Baby Learning'' (BL), to a standard deep learning setup (STDL). STDL uses AdamW with standard parameters (learning rate 1e-4, momentum b1 as 0.9), with weight updates after each batch, the same ViT-L model but with the popular ImageNet MAE checkpoint.  We implement the IID setting by sampling a sequence of random time steps (and associated target time steps) from random videos of the same base dataset. Note that while most approaches employ large batches, SGD with mini-batch size 1 is well known to work well~\cite{masters2018revisiting} and is even said to generalize better than using large batches. We match the number of frames that each method sees -- namely for batch size $>$ 1, the total number of learning steps is reduced proportionally.

\vspace{2mm}
\noindent \textbf{Dummy baseline.} For reference, we also include a dummy blind baseline, proposed in previous online learning works~\cite{hammoud2023rapid}, which exploits the temporal correlations in the target stream without seeing the input frames at all: this "model"  outputs the mean of the previously seen pixels at the same locations for depth and pixel, and the most frequent colors for segmentation, for the previous target. It produces random performance for out-of-stream evaluation where consecutive targets are unrelated\footnote{Another baseline that may be of interest is one that copies the previous input frames instead of target frames -- this could do well for both train and eval for pixels, but poorly for segmentation and depth}.

\vspace{2mm}
\noindent \textbf{Results} on all tasks are shown in Table~\ref{tab:main_table}.
We show visual results of a subset of these tasks in Fig.~\ref{fig:main_fig}. It is visible that our approach BL matches STDL with batch size 1 out-of-stream while outperforming it in-stream. In fig.~\ref{fig:performance_through_video} in the Appendix we also report results for STDL on a continuous stream, which does not work at all, and for BL on an IID stream which does best -- this shows that there are still many more improvements possible for learning from continuous streams.

\label{sec:experiments}

%% file: sec/5_conclusion.tex
\section{Conclusion}
\label{sec:conclusion}

We have presented a different perspective on continual learning by defining prediction tasks on a single very long stream of video and by proposing evaluations that measure both adaptation and generalization. We show that there is a permanent tension between the two aspects, but that it is possible to improve both by specializing the optimizer (lowering momentum, increasing the number of gradient aggregation steps between weight updates) and by pretraining the model appropriately.  Conversely, we show that just using Adam with default parameters performs poorly on single-stream learning, due to the large correlations in the data. This makes the proposed setting different from popular ImageNet-based continual learning setups where batches are employed, and the data is significantly uncorrelated - for example in class-by-class learning, seeing different web images of dogs will not break Adam the way consecutive frames in a video does.

We admit that the motivation for the research direction laid out in this paper may not be immediately crystal clear to everyone, particularly in the context of the current research landscape, that focuses on fitting larger and larger models to the whole internet. We are motivated by a possible future where people have their models in physical devices they carry around and train them via natural interaction -- by showing them the world from their perspective -- and teach them only the things that they believe the models should be taught, similar to the way we teach children. It is likely that such models, by not having to know everything, can be smaller, more efficient, have fewer internet biases and be more privacy-friendly.

\vspace{2mm}
\noindent \textbf{Acknowledgments.} We would like to thank Ross Goroshin and Ali Eslami for reviewing the paper and providing helpful comments, and Razvan Pascanu and Simon Osindero for insightful advice and interesting pointers into the literature and background work. 

%% file: sec/X_suppl.tex
\clearpage
\setcounter{page}{1}
\maketitlesupplementary

\section{Overview}
\label{sec:supp_overview}

This section provides additional results that we could not fit in the main paper, more details about the experiments, and extra visualizations. It also discusses two negative results we had along the way. See~\url{https://sites.google.com/view/one-stream-video} for  videos of future prediction pretraining and of ScanNet-stream training with the best models from table~\ref{tab:main_table}, in sequential and IID cases.

\subsection{Additional results}

\begin{figure*}
    \centering
    \includegraphics[width=\linewidth]{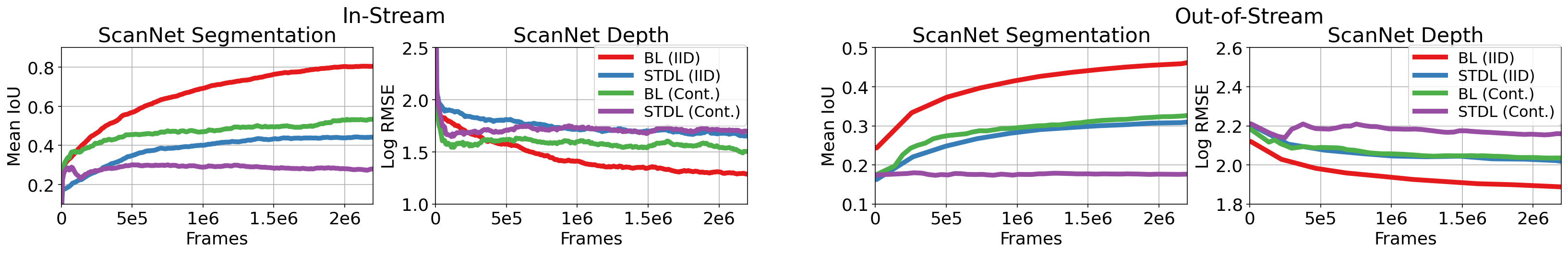}
    \caption{In-stream and out-of-stream performance on segmentation and depth estimation (with no offset in time) comparing a strong standard deep learning (STDL) approach, which uses an ImageNet-MAE checkpoint with batch size 1 and our approach (BL) when using a continuous video stream (Cont.) or IID data. STDL fails completely when fed a continuous stream. Note that here, pretraining from MAE is more closely aligned with the task than the video prediction pretraining objective we use, since displacement is 0 -- segmentation and depth align perfectly with the input frames. Yet, results suggest the video prediction objective leads to significant improvements. These results also suggest that there is a need for new continual learning techniques for bridging the gap between learning from IID data and a sequential stream (this becomes apparent when using the same strong pretrained model).}    \label{fig:iid_seq_tsd0}
\end{figure*}

\vspace{2mm}
\noindent \textbf{0 displacement results.} We show results for predicting segmentation and depth with no offset in time (the usual semantic segmentation and depth estimation tasks), comparing the strong \textit{standard deep learning} (STDL) approach to our approach (\textit{baby learning}, BL) on continuous streams and IID data (Fig.~\ref{fig:iid_seq_tsd0}). As expected, the standard deep learning approach performs very poorly on continuous streams. Our approach improves the performance in this setting and obtains outstanding performance in the IID setting, showing the effectiveness of pre-training via guided future prediction in both settings. Unless specified otherwise, plots are shown with exponential smoothing, with $\alpha=1e-3$, using Pandas~\cite{pandas_ewm}.

\begin{figure}
    \centering
    \includegraphics[width=.48\linewidth]{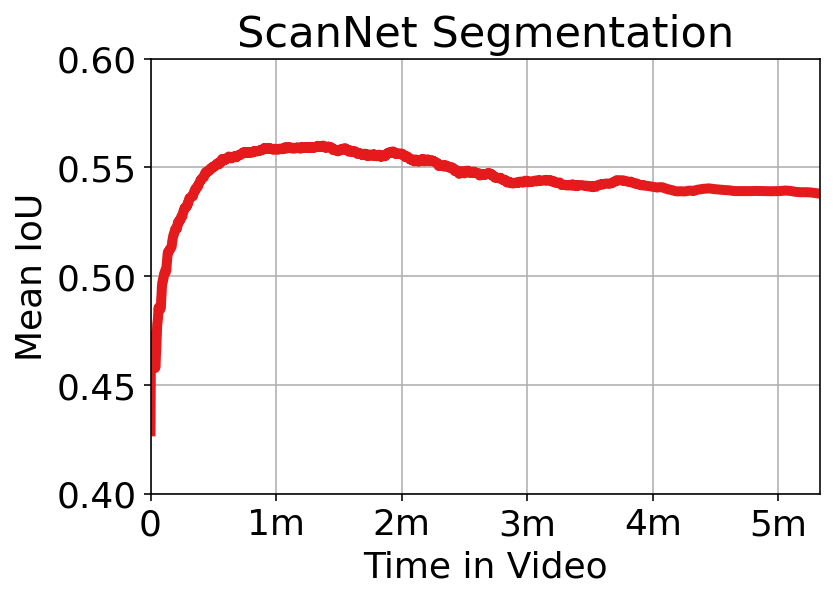}
    \includegraphics[width=.48\linewidth]{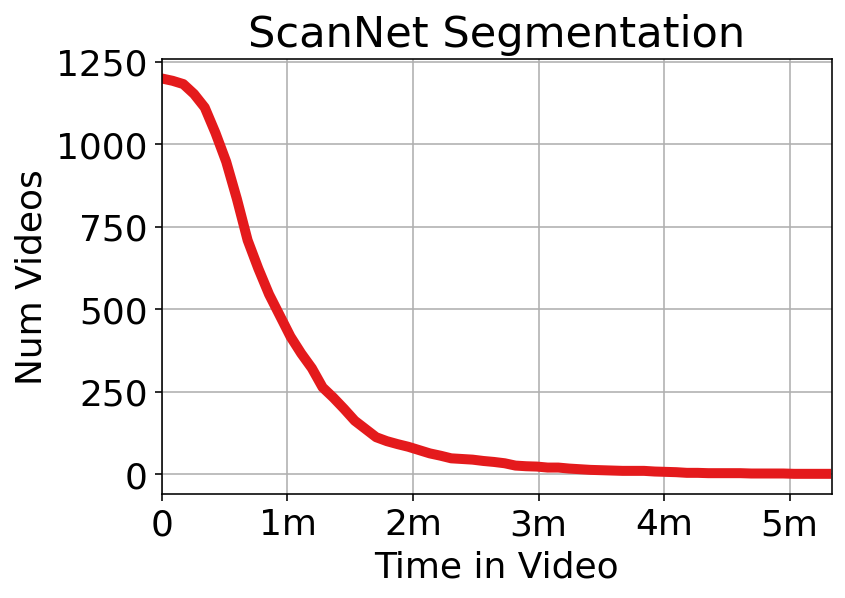}
    \caption{\textbf{Left}: In-stream Mean IoU plotted against time through the video averaged over all videos in 1 training run on ScanNet-stream  segmentation. \textbf{Right}: The number of videos that ScanNet-stream is composed of in training, with at least n minutes.}
    \label{fig:performance_through_video}
\end{figure}

\vspace{2mm}
\noindent \textbf{Performance over time in video.} Fig.\ref{fig:performance_through_video} shows the average in-stream performance over 1 training run on sequential data against the time spent in the video. This shows that the model performance improves rapidly over the first minute of each video it sees.

\begin{figure*}
    \centering
    \includegraphics[width=\linewidth]{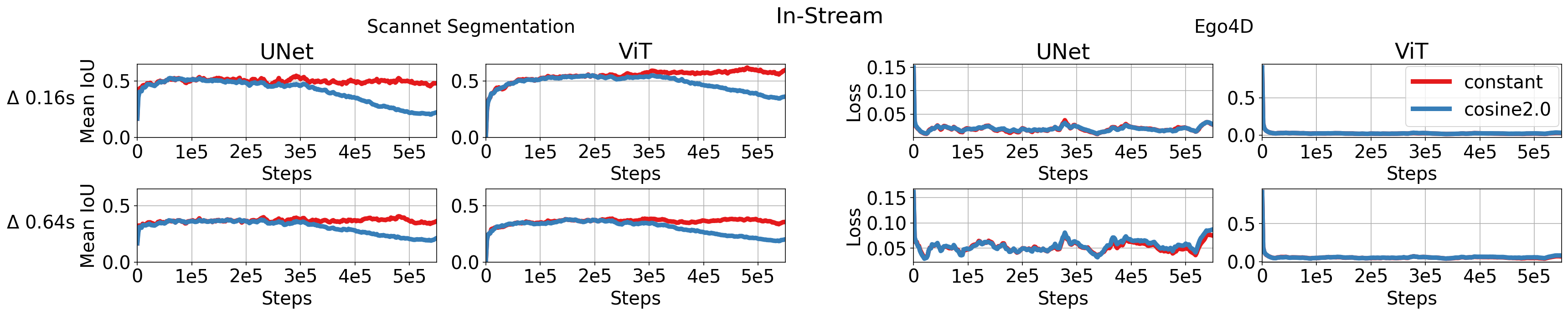}
    \caption{In-stream performance of 2 different learning rate schedules across 2 datasets, 2 displacements and 2 models. Our approach BL on a continuous stream is generally better than baseline STDL on an IID stream, for same batch size 1.}
    \label{fig:lr_separated_in_stream}
\end{figure*}

\begin{figure*}
    \centering
    \includegraphics[width=\linewidth]{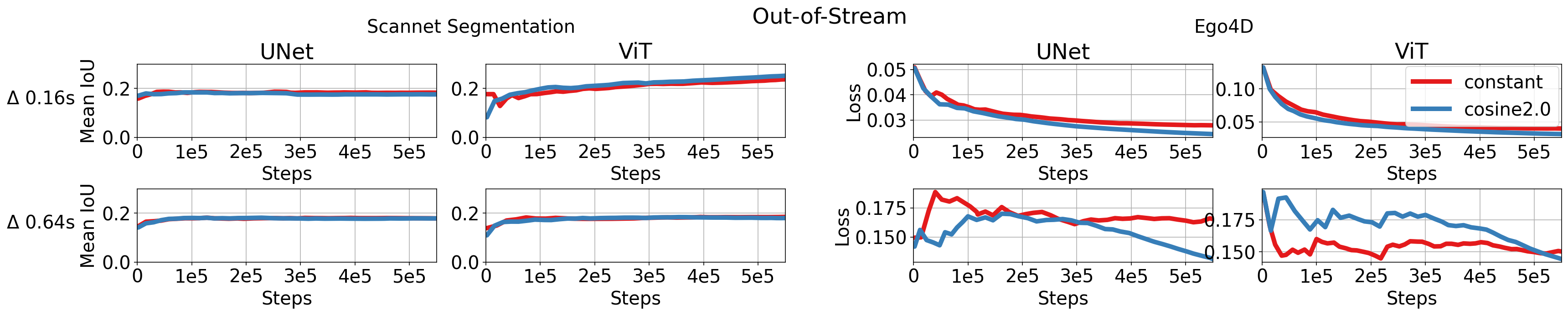}
    \caption{Out-of-stream performance of 2 different learning rate schedules across 2 datasets, 2 displacements and 2 models. Our approach BL on a continuous stream is competitive with STDL on an IID stream for the same batch size 1. It does fail badly for Ego4d with large displacement, but is better on ScanNet segmentation.}
    \label{fig:lr_separated_off_stream}
\end{figure*}

\begin{figure*}
    \centering
    \includegraphics[width=\linewidth]{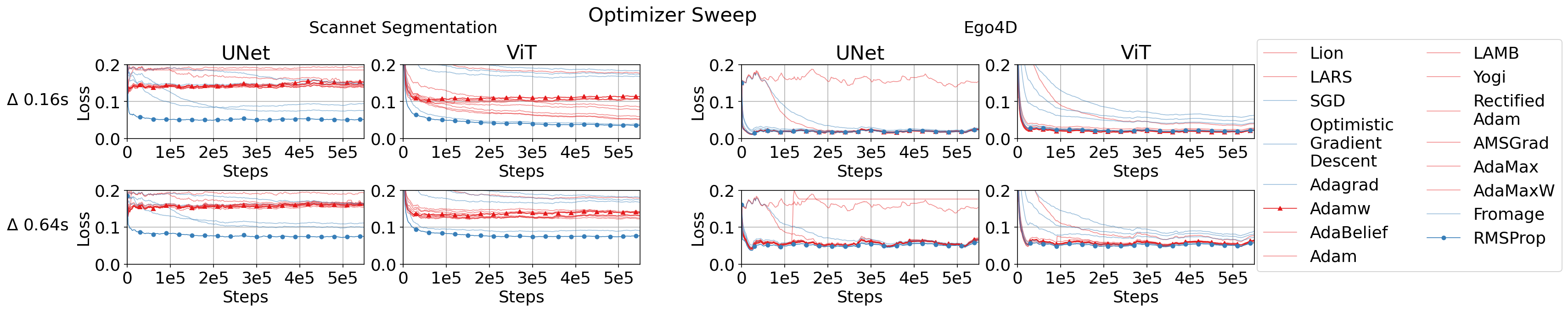}
    \caption{Training loss of different optimizers across 2 datasets, 2 displacements and 2 models.}
    \label{fig:optimizer_separated}
\end{figure*}

\begin{figure*}
    \centering
    \includegraphics[width=\linewidth]{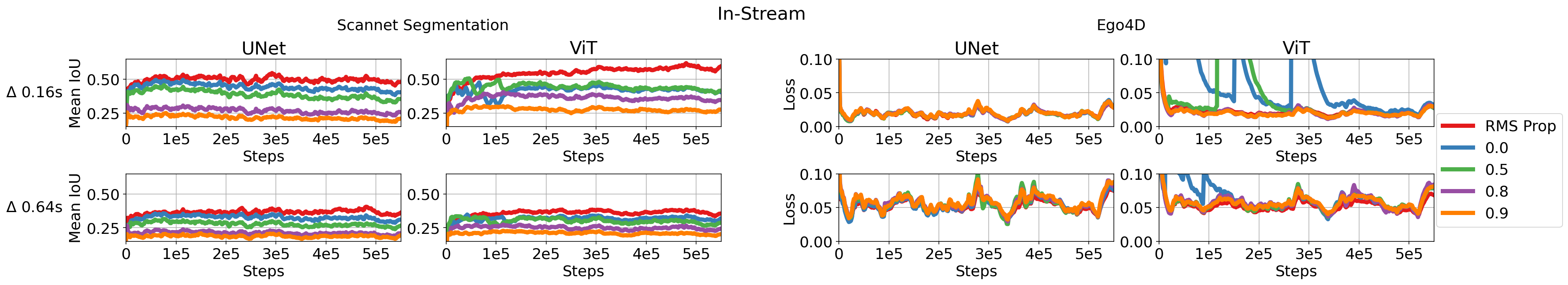}
    \caption{In-stream performance of RMSProp and AdamW with different levels of momentum across 2 datasets, 2 displacements and 2 models.}
    \label{fig:momentum_separated_in_stream}
\end{figure*}

\begin{figure*}
    \centering
    \includegraphics[width=\linewidth]{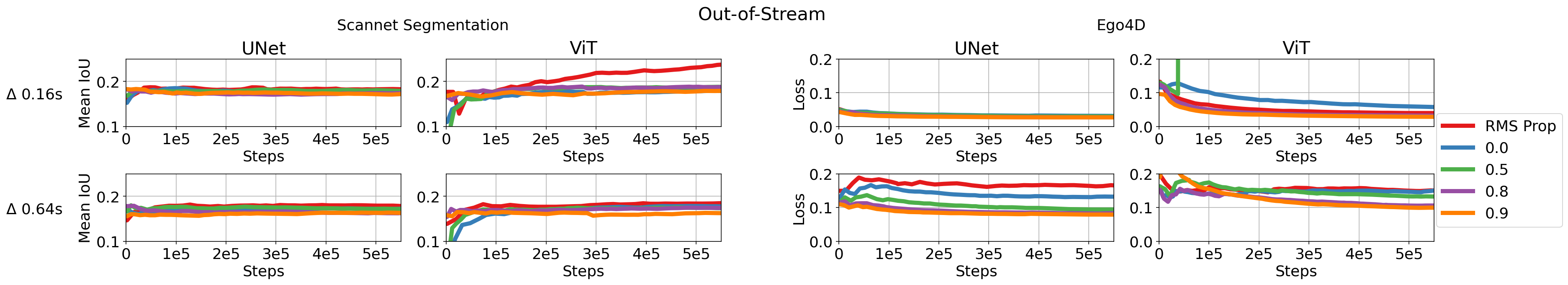}
    \caption{Out-of-stream performance of RMSProp and AdamW with different levels of momentum across 2 datasets, 2 displacements and 2 models.}
    \label{fig:momentum_separated_off_stream}
\end{figure*}

\vspace{2mm}
\noindent \textbf{Disaggregated results}. In the main body of the paper, many of our results are aggregated over multiple models, tasks and displacements so we can be more confident about their generality. Here we include disaggregated results for reference. Fig.~\ref{fig:lr_separated_in_stream} and Fig.~\ref{fig:lr_separated_off_stream} show the in-stream and out-of-stream evaluation for the 2 different learning rate schedules experimented with in Fig.\ref{fig:lr_sched_sweep} separately for different datasets, models, and displacements (these were aggregated over in the main paper). Fig.~\ref{fig:optimizer_separated} shows the training loss for the different optimizers experimented with in Fig.~\ref{fig:optimizer_sweep} separately for different datasets, models, and displacements. Fig.~\ref{fig:momentum_separated_in_stream} and Fig.~\ref{fig:momentum_separated_off_stream} show the in-stream and out-of-stream evaluation for RMS Prop and AdamW with different levels of momentum as in Fig.\ref{fig:momentum}, but  separately across datasets, models, and displacements.

\subsection{Experimental details}

\noindent \textbf{Pretraining.} We used the AdamW optimizer with learning rate 2e-4, weight decay of 0.05 and batch size 1024, updating the weights every training step. We did 1000 steps of linear warmup and afterwards kept the learning rate constant for 150k steps. We experimented with different temporal displacements for prediction, from 0.64s to 3.84s observing continuous improvement for all 3 objectives (vanilla, guided and masked future prediction). Guidance and masking are both implemented by splitting the input frames into a regular grid of non-overlapping patches and replacing a fixed percentage of them, respectively by patches from the future or by gray. We replaced 5\% and 10\% of the patches for guiding (5\% did better for longer displacements and overall) and 50\% or 75\% of the patches for masking (50\% did better for longer displacements and overall). 32x32 patches did best in both cases, compared to 16x16.

\vspace{2mm}
\noindent \textbf{Models.} Our ViT-L experiments used a standard model as described in \cite{dosovitskiy2020image} except that we inflated (replicated 4 times) the first layer when starting from ImageNet-pretrained checkpoints, to be able to deal with the inputs being 4 frames stacked along the channel dimension (unlike in training where ImageNet images had only 3 channels).

Our UNet experiments use a variant of the model described in~\cite{NEURIPS2020_4c5bcfec} processing at 4 resolutions, starting with 224x224. As in that paper, we use group norm throughout. At each resolution we use 8 residual blocks and 64 channels. At the lowest resolution, 28x28, we apply a self attention block with 4 heads.

\begin{figure}
    \centering
    \includegraphics[width=\linewidth]{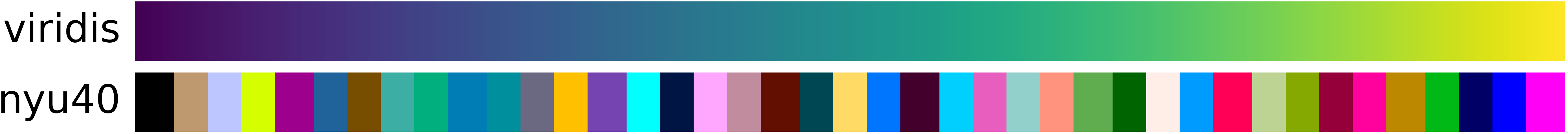}
    \caption{Colormaps that are used for mapping semantic labels and depth to RGB. Viridis is used for depth estimation task and NYU40 is used for segmentation.}
    \label{fig:colormaps}
\end{figure}

\noindent \textbf{Colormaps.} Our models output predictions for all tasks in RGB space. This is straightforward for the case of future frame prediction, but for some tasks  the outputs are originally defined in a different space. For example, for ScanNet semantic segmentation, outputs are traditionally represented by 40d one hot vectors. We map such spaces to RGB using the typical colormaps used by the datasets in their publications. For segmentation, we use the NYU40 colormap, which was used for visualizing the classes in ScanNet~\cite{dai2017scannet}. To evaluate, because predicted RGB values may not match exactly those corresponding to one label, we find the nearest neighbor color in the colormap, based on L2 distance and assign that pixel the corresponding label. For the depth task, we use a Viridis colormap from Matplotlib~\cite{Hunter:2007} which is a perceptually uniform sequential colormap with 256 colors and is commonly used to visualize depth in papers. We proceed similarly to the segmentation mapping to obtain the depth value from the rgb prediction. We assume a maximum depth of 8 and scale it to 1 when colormapping, then invert this normalization to find the predicted depth value. The colormaps can be found in Figure~\ref{fig:colormaps}.

\subsection{Additional visualizations}

\begin{figure}
    \centering
    \includegraphics[width=.48\linewidth]{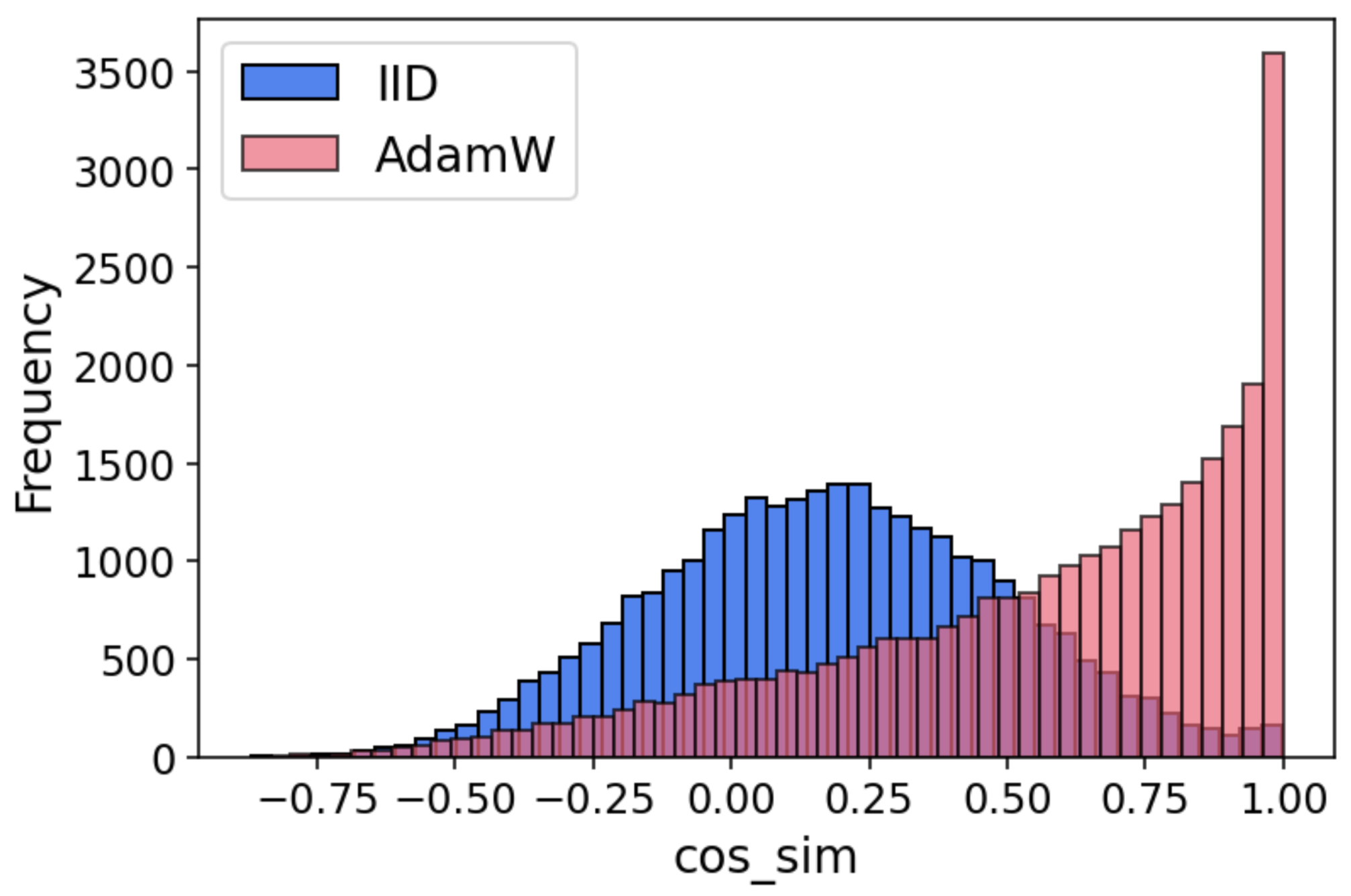}
     \includegraphics[width=.48\linewidth]{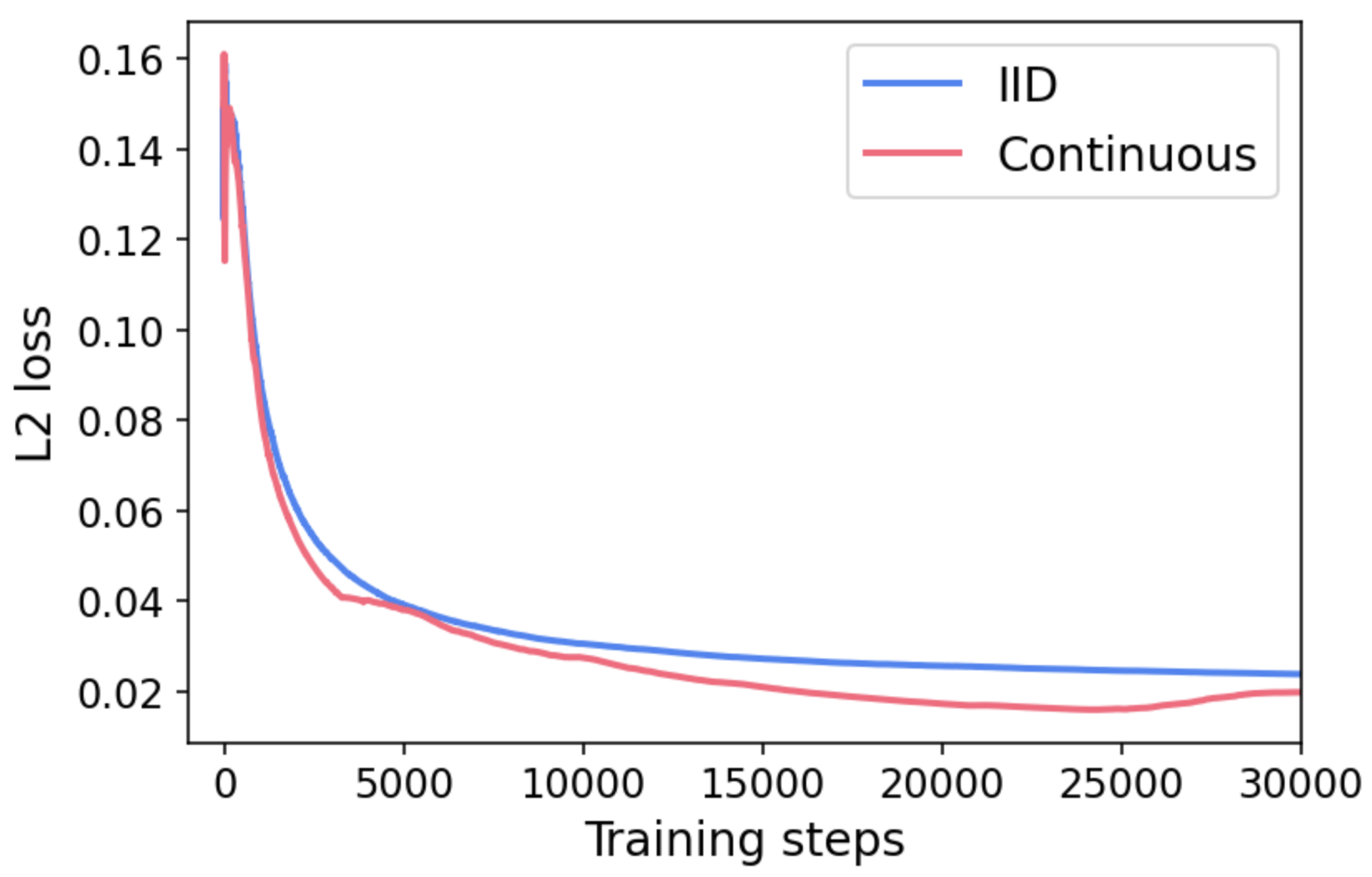}
    \caption{UNet training on Ego4D-stream for the future prediction task. \textbf{Left}: The cosine similarity of consecutive gradients is close to a normal distribution when training on IID data, but shows very strong correlations when training on a continuous video stream. \textbf{Right}: L2 loss.}
    \label{fig:cos-sim-ego4d}
\end{figure}

\noindent \textbf{Gradient correlations.} As showed in the main paper for the semantic segmentation task, Fig.~\ref{fig:cos_sim}, the cosine similarity between consecutive gradients shows very strong correlations when training from a continuous video stream as opposed to training with IID data. We observed the same trend when training on Ego4D for the task of future prediction, although in this case the loss seems less impacted by the correlations; see Fig.~\ref{fig:cos-sim-ego4d}.

\begin{figure*}
    \centering
    \includegraphics[width=\linewidth]{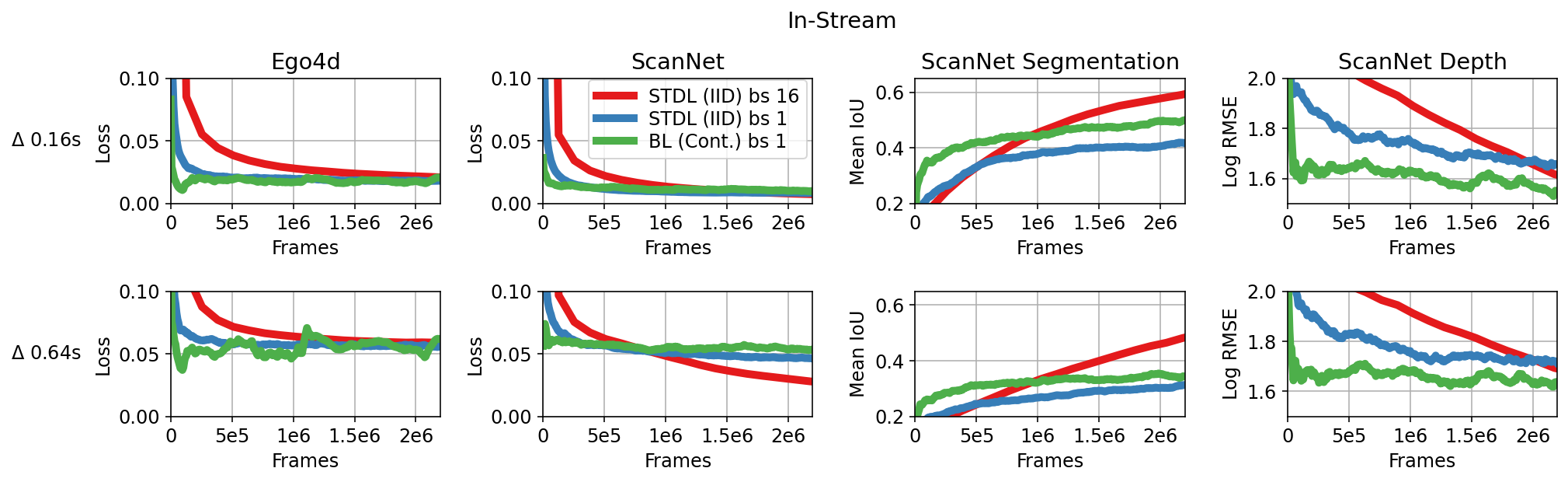}
    \caption{In-stream performance of a strong standard deep learning (STDL) approach with batch size 1 or 16 on IID data, and our approach (BL) when using a continuous video stream (Cont.).}
    \label{fig:in_stream_iid_seq}
\end{figure*}

\begin{figure*}
    \centering
    \includegraphics[width=\linewidth]{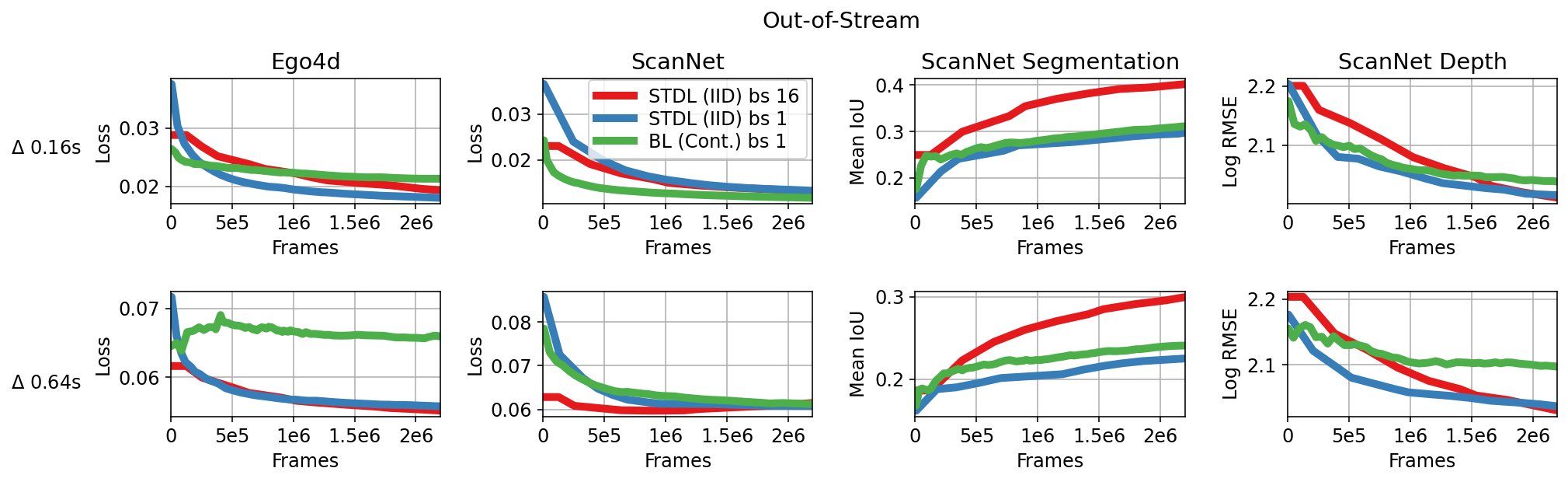}
    \caption{Out-of-stream performance of a strong  standard deep learning (STDL) approach with batch size 1 or 16 on IID data, and our approach (BL) when using a continuous video stream (Cont.).}
    \label{fig:off_stream_iid_seq}
\end{figure*}

\begin{figure*}
    \centering
    \includegraphics[width=\linewidth]{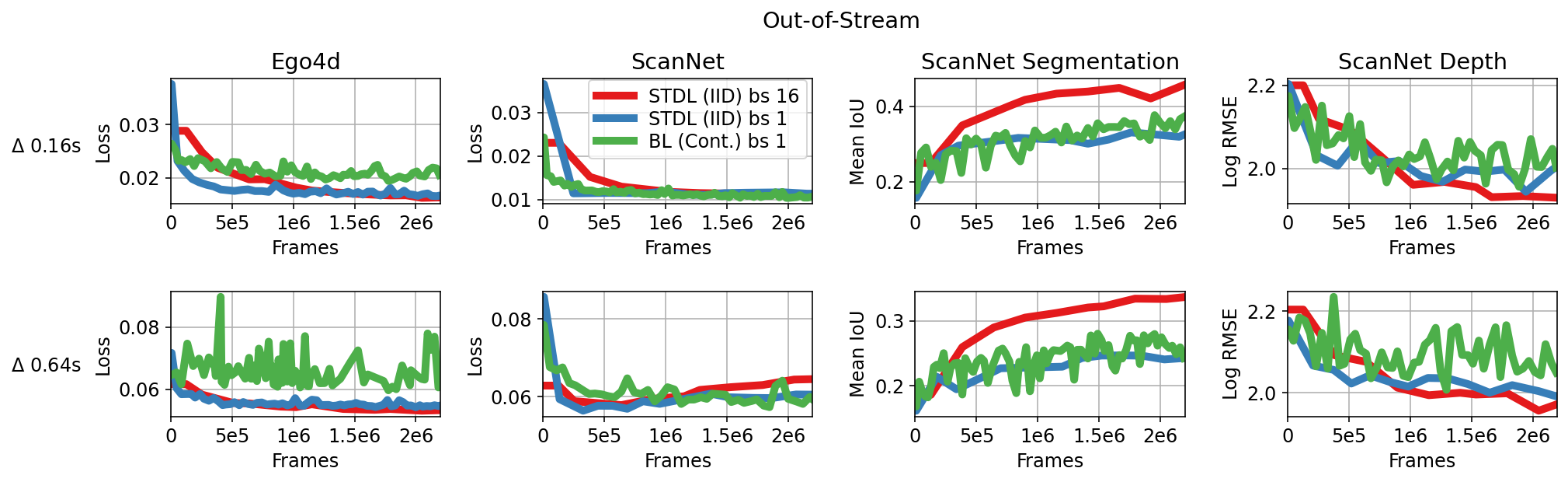}
    \caption{Same as Fig.\ref{fig:off_stream_iid_seq} but with no smoothing.}
    \label{fig:off_stream_iid_seq_nosmooth}
\end{figure*}

\vspace{2mm}
\noindent \textbf{In-stream / out-of-stream performance plots for table~\ref{tab:main_table}}. Fig~\ref{fig:in_stream_iid_seq} and Fig.~\ref{fig:off_stream_iid_seq} show the in-stream and out-of-stream evaluation curves corresponding to the results in Table \ref{tab:main_table} (Fig.~\ref{fig:off_stream_iid_seq_nosmooth} for version without any smoothing). 

\subsection{Negative results.}  

Other things we tried but were inconclusive or did not help:

\begin{itemize}
\item Elastic weight consolidation~\cite{Kirkpatrick2016OvercomingCF} is a well known continual learning technique that can be used in an online setting by creating a new copy of the model weights periodically, then penalizing certain departures away from these anchor weights afterwards. We implemented the simple L2 version of the method also described in the original paper.   While this helped for methods like Adam with default momentum, it did not provide additional benefit once we moved to RMSprop.
\item Augmentation -- random crops + flips per step. We experimented with data augmentation in an attempt to reduce deviation from IID, but did not observe advantages over a consistent augmentation for the whole video within a video stream. This was surprising and perhaps could be achieved by using more aggressive augmentation. We experimented with this only for semantic segmentation with time displacement of 0 (present prediction), as doing this for future prediction would require feeding in augmentation parameters to the model (otherwise the model could not possibly predict the target pixels, since the augmentation is random). 
\end{itemize}